\documentclass{article} 
\usepackage{colm2024_conference}

\usepackage{microtype}
\usepackage{hyperref}
\usepackage{url}
\usepackage{booktabs}
\usepackage{graphicx}
\usepackage{setspace}
\usepackage{caption}
\usepackage{amsmath}
\usepackage{booktabs} 
\usepackage{hyperref}
\usepackage[most]{tcolorbox}
\usepackage{times}
\usepackage{latexsym}
\usepackage{tabularx}
\usepackage{enumitem} 
\usepackage{multirow}
\newcommand*\samethanks[1][\value{footnote}]{\footnotemark[#1]}

\usepackage[T1]{fontenc}

\usepackage[utf8]{inputenc}

\usepackage{microtype}

\usepackage{inconsolata}
\usepackage{graphicx}
\usepackage{float} 
\usepackage{pifont}
\usepackage{array}

\title{AnyGPT: Unified Multimodal LLM with Discrete Sequence Modeling}

\author{Jun Zhan$^{1,}$\thanks{\ \  Equal contribution.}, 
Junqi Dai$^{1,}$\samethanks, 
Jiasheng Ye$^{1,}$\samethanks \\
{\bf Yunhua Zhou$^1$}, 
{\bf Dong Zhang$^1$}, 
{\bf Zhigeng Liu$^1$}, 
{\bf Xin Zhang$^1$} \\ 
{\bf Ruibin Yuan$^{2}$}, 
{\bf Ge Zhang$^{2}$}, 
{\bf Linyang Li$^1$}, 
{\bf Hang Yan$^3$}, 
{\bf Jie Fu$^{2}$} \\ 
{\bf Tao Gui$^1$}, 
{\bf Tianxiang Sun$^1$}, 
{\bf Yu-Gang Jiang$^1$}, 
{\bf Xipeng Qiu$^{1,}$\thanks{\ \  Corresponding author.}}\\ 
 $^1$ Fudan University\\
 $^2$ Multimodal Art Projection Research Community\\
 $^3$ Shanghai AI Laboratory
\\ \\
\texttt{\{jzhan22, jqdai22, jsye23\}@m.fudan.edu.cn} \quad
\texttt{xpqiu@fudan.edu.cn} \\ \\
\href{https://junzhan2000.github.io/AnyGPT.github.io/}{\texttt{https://junzhan2000.github.io/AnyGPT.github.io}}
}

\fancypagestyle{firstpage}{
  \lhead{\begin{picture}(0,0)\put(0,-6){\includegraphics[width=0.3\linewidth]{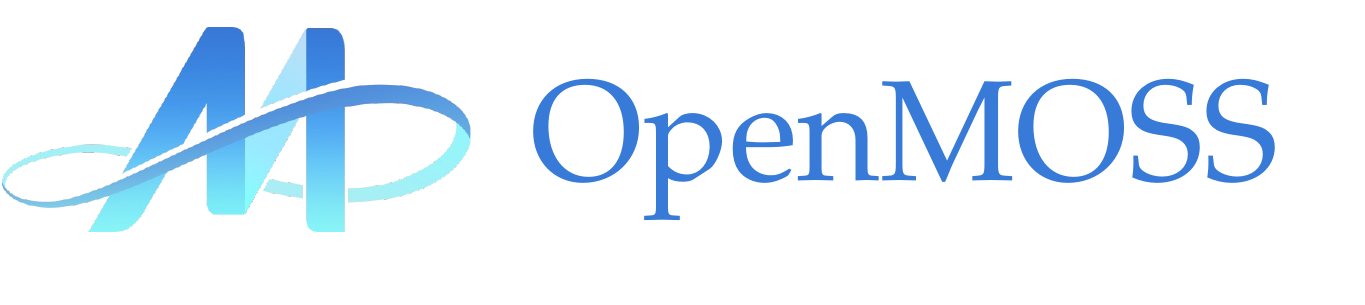}}\end{picture}}
}

\colmfinalcopy
\begin{document}

\maketitle
\thispagestyle{firstpage}

\begin{abstract}
We introduce AnyGPT, an any-to-any multimodal language model that utilizes discrete representations for the unified processing of various modalities, including speech, text, images, and music. 
AnyGPT can be trained stably without any alterations to the current large language model (LLM) architecture or training paradigms. Instead, it relies exclusively on data-level preprocessing, facilitating the seamless integration of new modalities into LLMs, akin to the incorporation of new languages.
We build a multimodal text-centric dataset for multimodal alignment pre-training. Utilizing generative models, we synthesize the first large-scale any-to-any multimodal instruction dataset. It consists of 108k samples of multi-turn conversations that intricately interweave various modalities, thus equipping the model to handle arbitrary combinations of multimodal inputs and outputs.
Experimental results demonstrate that AnyGPT is capable of facilitating any-to-any multimodal conversation while achieving performance comparable to specialized models across all modalities, proving that discrete representations can effectively and conveniently unify multiple modalities within a language model.
Demos are shown in \href{https://junzhan2000.github.io/AnyGPT.github.io/}{https://junzhan2000.github.io/AnyGPT.github.io/}.
\end{abstract}




\begin{figure*}[h] 
\centering
\includegraphics[width=0.83\textwidth]{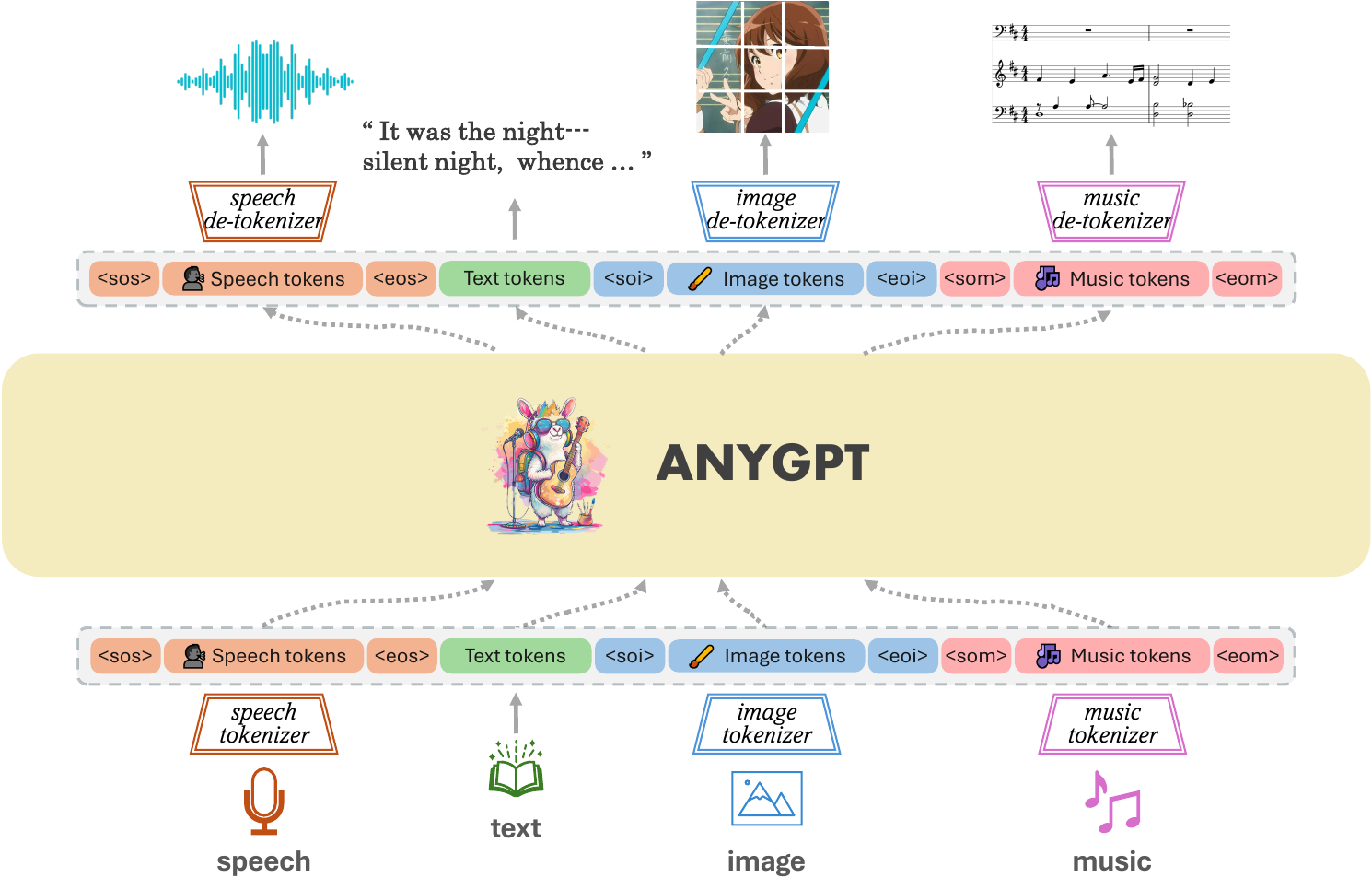}
\caption{An overview of the AnyGPT model architecture. All modalities are tokenized into discrete tokens, upon which the LLM performs multimodal understanding and generation autoregressively. Only data pre-processing and post-processing are required, with the model's architecture and training objectives remaining unaltered.}
\label{fig:model1}
\end{figure*}

\section{Introduction}

LLMs have exhibited remarkable proficiency in comprehending and generating human language. Nevertheless, their capabilities are confined to textual processing.
The real-world environment is inherently multimodal, with organisms perceiving and exchanging information through diverse channels, including vision, language, sound, and touch.

A promising objective in developing multimodal systems is to augment LLMs with the capacity for multimodal perception.
The dominant methodology involves the integration of multimodal encoders with the language model, thus empowering it to process information across various modalities and utilize its sophisticated text-processing abilities to produce coherent responses.
However, this strategy is limited to text generation and does not encompass multimodal output.

Pioneering efforts such as Emu~\citep{Sun2023GenerativePI}, SEED-LLaMA~~\citep{Ge2023MakingLS} and SpeechGPT~\citep{Zhang2023SpeechGPTEL} have made significant strides by enabling multimodal understanding and generation within language models. Yet, these models incorporate only a single non-textual modality, such as images or audio. While aligning text with one additional modality is relatively straightforward, integrating multiple modalities ($ N \geq 3 $) within a single framework—and achieving bidirectional alignment among them—poses a more formidable challenge.

Existing explorations in any-to-any multimodal generation have encountered obstacles: some~~\citep{Tang2023AnytoAnyGV} lacked a robust core language model, which impeded the system's reasoning and decision-making capabilities; Others, such as NExT-GPT~~\citep{wu2023next}, CoDi-2~~\citep{tang2023codi}, and Unified-IO2~~\citep{lu2023unified}, have employed separately pre-trained encoders and decoders. This approach results in representational inconsistencies between the inputs and outputs of the LLMs, which in turn complicates both training and inference processes. Moreover, stabilizing training with such diverse modalities necessitates substantial modifications to existing models and techniques.

To overcome these challenges,  we introduce AnyGPT, an any-to-any multimodal language model that employs discrete representations for unified processing.
AnyGPT is equipped with multimodal tokenizers that compress raw multimodal data, such as images and audio, into a sequence of discrete semantic tokens. 
These discrete representations enable the core LLM to unify tasks such as perception, understanding, reasoning, and generation in an autoregressive manner at the semantic level. 
Subsequently, de-tokenizers convert the discrete representations back into the original modal representations at the perceptual level.
Thanks to discrete representation, which filters out high-frequency, modality-specific perceptual information while preserving essential low-frequency semantic information~~\citep{ge2023planting, borsos2023audiolm, rombach2022high}, we can train our model stably without any alterations to the existing LLM architecture or training paradigms.
Instead, our approach relies solely on data-level preprocessing. This allows for the seamless integration of new modalities into LLMs, akin to the addition of new languages, and permits the direct application of existing LLM tools, which enhances the efficiency of both the training and inference stages.

Furthermore, to mitigate the scarcity of multimodal alignment data encompassing all modalities, we build a text-centric multimodal alignment dataset for pre-training.
Our goal is to use text as a bridge, by aligning other modalities with text, to achieve mutual alignment among all modalities, since natural language is the most refined modality of semantic representation and is present in the majority of multimodal alignment datasets.
To endow the model with the capability to comprehend and generate content interwoven with multiple modalities, we employ advanced generative models to synthesize a multimodal instruction dataset, AnyInstruct-108k. This dataset, comprising 108k samples of multi-turn conversations, enables AnyGPT to handle arbitrary combinations of multimodal inputs and outputs.

Experimental results demonstrate that AnyGPT achieves zero-shot performance comparable to that of specialized models across various modalities. Furthermore, extensive case studies corroborate AnyGPT's remarkable ability to facilitate any-to-any multimodal dialogue, substantiating the feasibility of using discrete representations to unify multiple modalities.

Our contributions include the following:

\begin{itemize}
  \item We propose AnyGPT, a token-based any-to-any multimodal language model which can understand and generate various modalities, including speech, text, images, and music.
  \item One key challenge is the lack of multimodal interleaved instruction-following data. 
  We develop a pipeline using generative models to build AnyInstruct-108k, a dataset comprising 108k multi-turn dialogues with interleaved multimodal elements.
  \item We demonstrate discrete representations can effectively unify multiple modalities within a language model. 
\end{itemize}

\section{Related Work}

\subsection{Multimodal Large Language Models}
To enable cross-modal perception in LLM, a common approach is to connect pre-trained encoders of other modalities as adaptors. However, these models are often limited to text generation.

To empower LLMs with multimodal generation capabilities, ~~\citet{Tang2023AnytoAnyGV} introduces a frozen text-to-image diffusion model and learns the mapping between the LLM's embeddings and the diffusion model. ~\citet{Sun2023GenerativeMM} utilizes continuous embeddings to represent the image, calculating either a  loss for the next token prediction or   the next visual embedding regression.    In contrast, SEED-LLaMA~~\citep{Ge2023MakingLS} trains an image discretization tokenizer to encode the original image  into discrete tokens. Through a unified next token prediction task, it achieves unified image understanding and generation. Similarly, in the field of speech, SpeechGPT~~\citep{Zhang2023SpeechGPTEL} enables LLMs to have inherent cross-modal conversation capabilities through discrete speech representation. VideoPoet~~\citep{kondratyuk2023videopoet} employs a decoder-only transformer architecture that processes multimodal inputs – including images, videos, text, and audio, and is capable of generating videos and audio.

To achieve multimodal generation across various modalities on LLMs, NExT-GPT~~\citep{wu2023next} utilizes existing high-performance encoders and decoders, connected by a small number of projection layer parameters. 
However, NExT-GPT does not train the LLM, which may result in suboptimal performance. Moreover, its representation of multimodal input and output lacks a unified form, which poses challenges in unified training and inference.

\subsection{Multimodal Discretization}
To create a unified multimodal language model, a common approach is to use discretization. A typical method is VQ-VAE~~\citep{van2017neural}. This involves maximizing the restoration of the original representation from the compressed tokens. Some  studies~~\citep{Defossez2022HighFN,Zeghidour2021SoundStreamAE}  incorporate residual quantization mechanisms to further enhance fidelity.

In addition to VQ-VAE based tokenizers, some tokenizers focus on extracting high-level semantic representations.~~\citet{Ge2023MakingLS}  discretizes the image into semantic-level.
The SpeechTokenizer~~\citep{zhang2023speechtokenizer}, based on the RVQ-VAE structure, enables the first layer of tokens to retain the semantic information of speech, and the remaining layers to supplement residual information information, achieving a disentanglement of semantic and acoustic information.

\section{AnyGPT }
\label{qwer}
Our interest lies in facilitating the generation of any modality to any modality with LLMs. 
To realize this, we propose a comprehensive framework that can be uniformly trained.
As illustrated in Figure \ref{fig:model1}, this framework is composed of three main components: (${1}$) multimodal tokenizers, (${2}$) a multimodal language model serving as the backbone, and (${3}$) multimodal de-tokenizers. 
The tokenizers transform continuous non-text modalities into discrete tokens, which are subsequently arranged into a multimodal interleaved sequence. 
Then the sequences are trained by the language model using the next token prediction training objective. 
During the inference process, multimodal tokens are decoded back into their original representations by the associated de-tokenizers.
To enrich the quality of generation, multimodal enhancement modules can be deployed to post-process the generated results, including applications like voice cloning or image super-resolution. 
In the following section, we will introduce the details of each module.

\subsection{Tokenization}\label{sec:tokenization}
\begin{table}[h!]
\setlength{\tabcolsep}{1pt}
\centering\small
\begin{tabular}{l|ccc}
\toprule
Modality & Image & Speech   & Music \\
\midrule
Vocab Size        & 8192  & 1024              & 4096           \\
Tokens per Sample & 32 / per image  & 50 / s            & 200 / s        \\
RVQ               & \ding{56}              & \ding{52}                & \ding{52}              \\
Input Size        & 224*224        & variable duration & 5s       \\
\bottomrule
\end{tabular}
\caption{Details of tokenization of different modalities.}
\label{tab:tokens}
\end{table}
\paragraph{Image Tokenizer}
We utilize the SEED tokenizer~~\citep{ge2023planting} for image tokenization. The SEED tokenizer consists of several components, including a ViT encoder ~~\citep{dosovitskiy2020image}, Causal Q-Former, VQ Codebook ~~\citep{van2017neural}, multi-layer perception (MLP), and a UNet decoder ~~\citep{ronneberger2015u}.
SEED takes a $224 \times 224$ RGB image as input, and the ViT encoder encodes the image into $16 \times 16$ patches, then the Causal Q-Former converts the  patch features into 32 causal embeddings. A codebook with 8192 entries discretizes the embeddings into a sequence of quantized codes. An MLP is employed to decode the visual codes into a generation embedding, which is aligned with the latent space of the pre-trained unCLIP Stable Diffusion(unCLIP-SD)~~\citep{rombach2022high}. Finally, the UNet decoder is used to restore the generation embedding to the original image.

\paragraph{Speech Tokenizer}
The tokenizer for speech we utilize is SpeechTokenizer~~\citep{zhang2023speechtokenizer}, adopting an encoder-decoder architecture with residual vector quantization (RVQ).
The SpeechTokenizer compresses single-channel audio sequences into a discretized matrix using eight hierarchical quantizers, each with 1,024 entries, and achieves a frame rate of 50 Hz.
The first quantizer layer captures semantic content, while layers 2 to 8 encode paralinguistic details. A 10-second audio is thus transformed into a $500 \times 8$ matrix, splitting into semantic and acoustic tokens. We adopt a SpeechTokenizer variant pre-trained on the Commonvoice~~\citep{ardila2019common} and Librispeech~~\citep{panayotov2015librispeech} datasets.

In AnyGPT, the Large Language Model (LLM) is employed to model the semantic tokens, while a voice cloning model supplements the remaining paralinguistic information. As a result, the size of the voice vocabulary in the LLM is equivalent to the size of one codebook, which is 1024. Further details will be discussed on in Section ~\ref{sec:multimodal generation}.

\paragraph{Music Tokenizer}

Although speech and music share similar data formats, their substantial content differences lead us to treat them as distinct modalities, each equipped with its own tokenizer. 
For music, we employ Encodec~~\citep{Defossez2022HighFN}, a convolutional auto-encoder with a latent space quantized using Residual Vector Quantization (RVQ), as the music tokenizer. We use an available off-the-shelf variant of the Encodec\footnote{\url{https://huggingface.co/facebook/encodec_32khz}}  pre-trained on 20k pieces of music tracks. This variant processes 32 kHz monophonic audio, and achieves a frame rate of 50 Hz. The embeddings generated are quantized using an RVQ with four quantizers, each with a codebook size of 2048, resulting in a combined music vocabulary size of 8192.

We encode 5 seconds music into 250 latent frames, ultimately generating a $250 \times 4$ codes matrix. To enable the language model predict entire music clip, we flatten the 4-layer music codes into a causal sequence in a frame-by-frame manner. The language model begins by predicting the initial four tokens of the first frame and continues in a similar fashion for the subsequent frames.

\subsection{Language Model Backbone}

\paragraph{Expanding vocabulary}
To incorporate multimodal discrete representations into pre-trained LLMs, we expand the vocabulary with new modality-specific tokens, and consequently extend the corresponding embeddings and prediction layer, the newly incorporated parameters are initialized randomly. 
The tokens from all modalities combine to form a new vocabulary, where each modality is trained within the language model to align in a shared representational space. 
The size of this enhanced vocabulary, denoted by $V$, is the summation of the vocabulary sizes across all modalities, that is, $V = \sum_{i=1}^{n} V_i$, where $V_i$ signifies the vocabulary size of the $i$-th modality. 

\paragraph{Unified Multimodal Language Model}
Equipped with the modality-specific tokenizers, we can compress multimodal data into discrete token sequences, which can be trained by the language model using the next token prediction loss. 
This naturally enables the core LLM to unify tasks such as perception, understanding, reasoning, and generation in an autoregressive manner.

We employ the LLaMA-2~~\citep{touvron2023llama} 7B as the backbone, which is pre-trained on 2 TB of text tokens. Apart from reshaping the embedding matrix and prediction layer, the rest of the language model remains unaltered.

\subsection{Multimodal Generation }\label{sec:multimodal generation}
The generation of high-quality multimodal data, including high-definition images,  and high-fidelity audio, presents a substantial challenge. These data typically necessitate a large number of bits for accurate representation, resulting in long sequences which is particularly demanding for language models, as the computational complexity increases exponentially with the length of the sequence.

To tackle this, we adopt a two-stage framework for high-fidelity generation, comprising semantic information modeling and perceptual information modeling. 
First, the language model is tasked with generating content that has undergone fusion and alignment at the semantic level. Then, non-autoregressive models convert multimodal semantic tokens into high-fidelity multimodal content at the perceptual level, striking a balance between performance and efficiency.

Specifically, we employ SEED tokens, aligned with the diffusion latent space, for visual language modeling. 
 Semantic-level SEED tokens are decoded into high-quality images by a Diffusion Model, which is renowned for its superior generation capabilities. 
For speech, we utilize SoundStorm~~\citep{borsos2023soundstorm}, a non-autoregressive Masked Language Model, trained to generate SpeechTokenizer's acoustic tokens from semantic tokens. 
We train a variant of Soundstorm, which is trained using the SpeechTokenizer on the Multilingual LibriSpeech(MLS) dataset~\citep{pratap2020mls}.
Subsequently, the SpeechTokenizer's decoder transforms all speech tokens into raw audio data. 
This approach enables AnyGPT replicate the voice of any speaker using a 3-second speech prompt, while significantly reducing the length of the voice sequence for LLM.
For music, we employ Encodec tokens to filter out high-frequency details beyond human perception, and then use the Encodec decoder to reconstruct these tokens into high-fidelity audio data.

\section{Multimodal Data }
\subsection{Pre-training Data }

To enable the generation from any modality to any other, it is crucial to have data that is well-aligned across these modalities. 
Unfortunately, such data is notably scarce. 
To address this challenge, we build a text-centric bi-modal alignment dataset. Here, text is employed as a vital intermediary to bridge the gap between various modalities. By aligning different modalities with the textual modality within a language model, we aim to achieve mutual alignment amongst all modalities.

\begin{figure}[H] 
  \centering
  \includegraphics[width=0.45\textwidth]{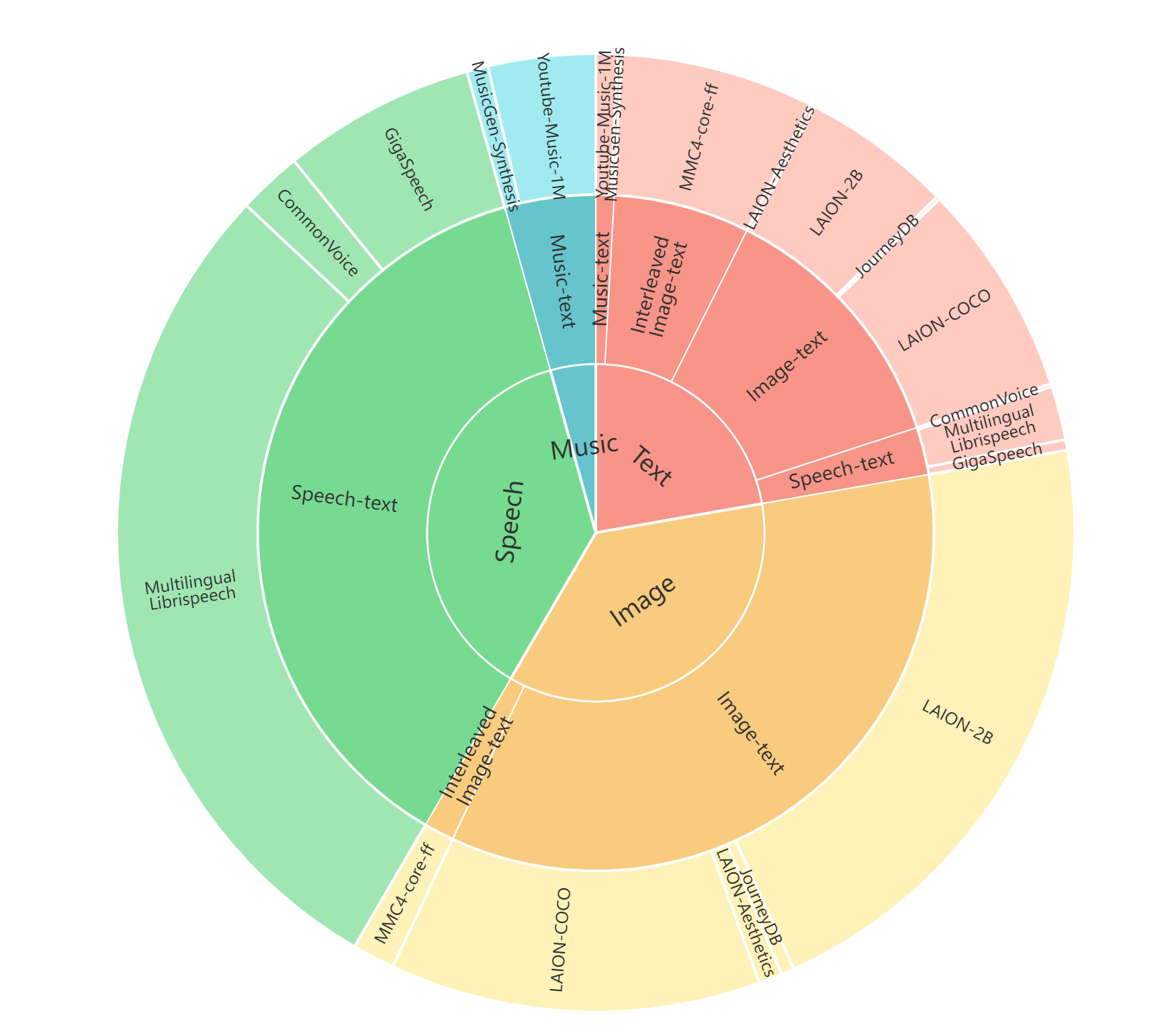}
  \caption{Pre-training data distribution, segmented by token counts, with the inner section indicating the modality, the middle detailing data types, and the outer specifying individual datasets.}
  \label{fig:pre-training data}
  \end{figure}

The representational forms and types of information vary greatly across different modalities, To facilitate a standardized comparison of data volumes across various modalities, we have adopted a quantification approach based on token counts.
Figure \ref{fig:pre-training data} presents all the data used in pre-training and their respective proportions. 
A certain level of oversampling is applied to modalities with comparatively lower data quantities, to attain a balanced representation of diverse data types within a single batch. 
More details are in Appendix \ref{sec:pre-training Data}.

\paragraph{Image \& Text}

We utilized image-text pairs from LAION-2B~~\citep{schuhmann2022laion}, LAION-COCO~~\citep{laioncoco}, LAION-Aesthetics~~\citep{laion-aesthetics} and JouneyDB~~\citep{pan2023journeydb}. LAION-2B provides images paired with noisy alt-texts sourced from the web, while LAION-COCO represents a 600M subset of this, captioned by BLIP. We refined these datasets by filtering for text quality, image aspect ratio, and clip score, etc., yielding a high-quality corpus of 300M pairs. To enhance the overall image generation fidelity, we supplement our data with the high-quality LAION-Aesthetics subset and the synthetic dataset JourneyDB from Midjourney.

We also incorporate image-text interleaved data to adapt the model to an interleaved mode. We deploy the Multimodal-C4 (MMC4) dataset~~\citep{zhu2023multimodal}, an enhanced version of the text-only C4~~\citep{raffel2020exploring}. Specifically, we utilize the MMC4-core split, consisting of  7.3M documents. 

\paragraph{Speech \& Text}

We collect several large-scale English Automatic Speech Recognition (ASR) datasets, including Gigaspeech~~\citep{chen2021gigaspeech}, Common Voice~\citep{ardila2019common}, and Multilingual LibriSpeech(MLS)~\citep{pratap2020mls}.
These datasets are sourced respectively from online platforms, volunteer crowdsourcing, and audiobooks, collectively constituting a corpus of 57,000 hours of speech-text pairs, encompassing a wide variety of speakers, domains, and recording environments.

\paragraph{Music\&Text}
We embark on an extensive data collection process by crawling over one million music videos from the Internet. The core step involves matching the titles of these videos with corresponding songs using the Spotify API. Subsequently, we harvest a comprehensive set of metadata for each music audio, including video titles, descriptions, keywords, playlist names, and Spotify lyrics. This metadata is formatted into JSON and fed into GPT-4~\citep{achiam2023gpt} for processing. GPT-4's role is pivotal as an intelligent caption generator; it utilizes the noisy metadata to extract meaningful information and succinctly summarize it into coherent sentences. This approach allows us to generate high-quality text captions for a large amount of music audio, effectively minimizing the occurrence of hallucinations in the dataset.

\paragraph{Training Sample Construction.}

To train the Language Model (LM), we employ various templates to construct multimodal sentences, which the LM then processes autoregressively. Further training details can be found in Appendix \ref{sec:Training Sample Construction}.
Additionally, We observe significant variation in sentence lengths across different modalities and datasets. To enhance training efficiency, samples from the same dataset are concatenated into a long sequence, adhering to the model's maximum sequence length. Consequently, each token in the sequence contributes to the loss.

\subsection{Multimodal Interleaved Instruction Data Construction }
\begin{figure*} \centering \includegraphics[width=0.95\textwidth]{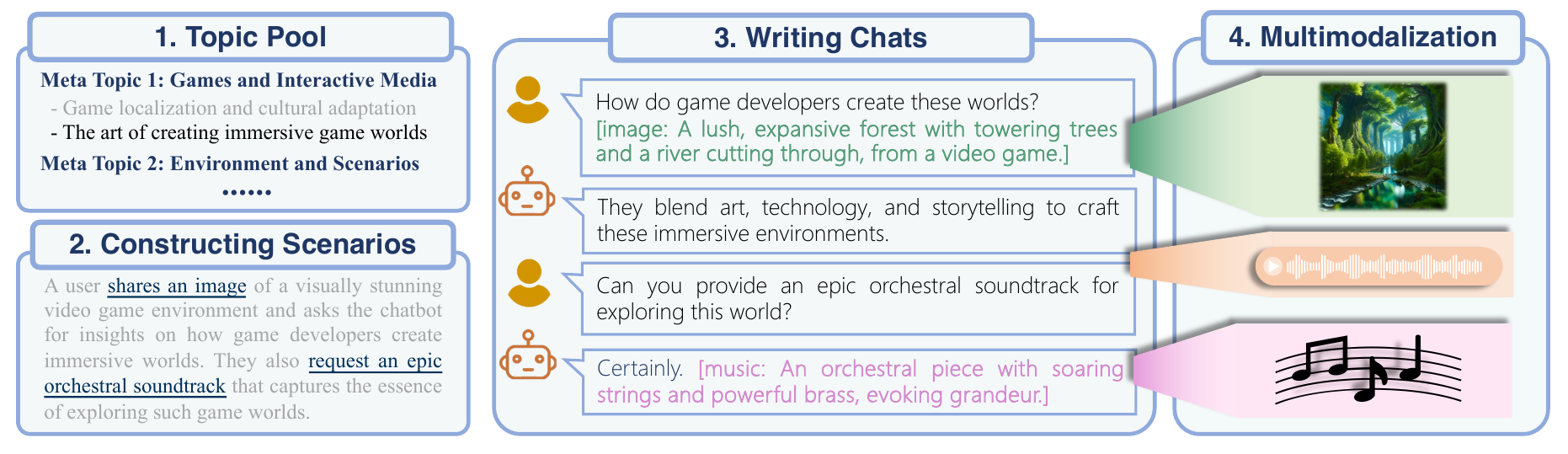} 
\caption{The construction process of the multimodal interleaved instruction dataset AnyInstruct is divided into two stages: Generation of text-based conversations incorporating multimodal elements and Text-to-Multimodality Conversion. The first stage generates topics, scenarios, and textual dialogues, while the second stage produces the final multimodal dialogues.} 
\label{fig:data synthesis} 
\end{figure*}

\label{sec: mm-interleaved-data}

Effective human-machine interaction should permit the exchange of information in a variety of interleaved modalities. However, the increasing number of modalities in conversation significantly complicates the data collection process. To our knowledge, there is currently no large-scale instruction dataset involving more than two modalities. This poses a significant limitation on the development of a comprehensive model capable of managing dialogues with multiple, intertwined modalities.

To overcome this limitation, we draw inspiration from the most recent research on data synthesis~~\citep{wang2022self, wu2023next}, and build a dataset comprised of 108k multi-turn conversation samples with generative models. With careful curation, each synthetic conversation integrates multiple modalities—text, speech, images, and music—in an interleaved manner. Specifically, our data synthesis process is carried out in two stages, as illustrated in Figure ~\ref{fig:data synthesis}.

\paragraph{Generation of text-based conversations incorporating multimodal elements.}
In this phase, we employ GPT-4 to generate a series of text-based conversations. Notably, we incorporate non-text modality in the form of their textual descriptions within these conversations. To ensure high-quality data at scale, we divide this stage into three steps.
\textbf{(1)}~Initially, we brainstorm 100 meta topics to cover a broad spectrum of scenarios related to audiovisual elements and we employ GPT-4 to expand these meta-topics into 20,000 specific topics.
\textbf{(2)}~Subsequently, we prompt LLM to generate specific dialogue scenarios based on these topics. 
Acknowledging the intrinsic constraints of a text-based LLM in generating multimodal elements, we prepare several demonstrations that encompass as many modality combinations as possible. While generating scenarios, a subset is sampled from this demonstration pool, serving as examples for the LLM. This approach guides the model to effectively synthesize varied and contextually appropriate conversational scenarios.
\textbf{(3)}~Finally, we utilize GPT-4 to generate multi-turn conversations derived from scenarios. 
In these synthesized dialogues, multimodal elements, including images and music, are depicted through detailed textual representations. 
We curate a diverse range of conversation examples, similar to scenario generation, to prompt the model into creating dialogues with the widest possible variety of modalities.
As a result, we compiled a substantial corpus of multimodal conversational data in solely textual format.

\paragraph{Text-to-Multimodality Conversion.}
In this phase, we employ advanced generative models to convert textual descriptions into multimodal elements. We use OpenAI's DALL-E-3~\citep{betker2023improving} for image generation, MusicGen~\citep{copet2023simple} for music composition, and Microsoft Azure's text-to-speech API~~\citep{Azure_text2speech} for speech synthesis from user's instructions and model's text responses.

After filtering, we obtain a dataset of 108k high-quality multimodal dialogues, featuring a variety of multimodal combinations. This dataset includes around 205k images, 503k voice recordings, and 113k music tracks. Additionally, we enhanced our dataset by extracting dialogues from existing text-only instruction datasets well-suited for spoken narration. This results in 100k voice dialogues through the employment of text-to-speech models.

The two-stage approach efficiently collected a diverse array of high-quality multimodal conversations at scale. Appendix \ref{sec:anyinstrurct-prompts-appendix} provides the prompts used during the data synthesis process.

\section{Experiment }

\subsection{Evaluation }
We evaluate the fundamental capabilities of the pre-trained base AnyGPT (Section~\ref{qwer}), covering   multimodal understanding and generation tasks for all modalities. This evaluation aimed to test the alignment between different modalities during the   pre-training process. Specifically,  we test both text-to-X and X-to-text tasks for each modality,  where X is image, music, and speech separately.   

To simulate real-world scenarios, all evaluations are conducted in a \textit{zero-shot} mode. This means that AnyGPT will be \textit{not fine-tuned nor pre-trained} on downstream training samples during the evaluation process. This challenging evaluation setting requires the model to generalize to an unknown test distribution, showcasing the generalist abilities of AnyGPT across different modalities. The evaluation results demonstrate that AnyGPT, as a generalist multimodal language model, achieves commendable performance on various multimodal understanding and generation tasks.
\subsubsection{Image}

\paragraph{Image Understanding}We assess the image comprehension 
capabilities of AnyGPT on the image captioning task.  The comparison results are presented in Table~\ref{tab:coco}.   We utilize the MS-COCO 2014 captioning benchmark~~\citep{Lin2014MicrosoftCC} and adopt the \textit{Karpathy split testset} following previous studies~~\citep{Li2023BLIP2BL, Tang2023AnytoAnyGV}.   

\begin{table}[h!]
\centering\small
\setlength{\tabcolsep}{5pt}
\begin{tabular}{p{5cm}p{2cm}<{\centering}}
\toprule
Method    & CIDEr  $\uparrow$  \\
\midrule
Flamingo (9B)~~\citep{Alayrac2022FlamingoAV}     & 79.4 \\
Flamingo (80B)       & 84.3 \\
Emu (14B)~~\citep{Sun2023GenerativePI}              & 112.4        \\
DreamLLM (8B)~~\citep{Dong2023DreamLLMSM}   & 115.4 \\
InstructBLIP (14B) ~~\citep{Dai2023InstructBLIPTG}  &{\color[HTML]{9B9B9B} 102.2}\\
SEED-LLaMA (8B)~~\citep{Ge2023MakingLS}      & \textbf{{\color[HTML]{9B9B9B} 123.6}}  \\
\midrule
AnyGPT (8B)       & 107.5  \\
\bottomrule
\end{tabular}
\caption{Comparison results on image captioning task. Results in {\color[HTML]{9B9B9B} grey} indicate models have trained on training samples. }
\label{tab:coco}
\end{table}
\paragraph{Image Generation}  The results of the text-to-image generation task are presented in Table~\ref{tab:t2i}.  To ensure consistency with previous research~~\citep{Koh2023GeneratingIW,Ge2023MakingLS,Sun2023GenerativeMM}, we randomly select 30k images from the MS-COCO validation set and use  CLIP$_{score}$ as the evaluation criterion. This metric computes a similarity score between the generated image and  its corresponding caption from a real image, based on CLIP-ViT-L~~\citep{Radford2021LearningTV}.

\begin{table}[]
\centering\small
\begin{tabular}{lc}
\toprule
Method     & CLIP$_{score}$  $\uparrow$ \\
\midrule
GILL~~\citep{Koh2023GeneratingIW}      & 0.67   \\
Emu       & 0.66   \\
SEED-LLaMA & \textbf{0.69}   \\
\midrule
AnyGPT     & 0.65  \\
\bottomrule
\end{tabular}
\caption{Comparison results on  text-to-image generation task. We adopt MS-COCO captions to generate images and calculate the CLIP similarity score (CLIP$_{score}$)  for evaluation.}
\label{tab:t2i}
\end{table}

\subsubsection{Speech }

\paragraph{ASR}We evaluate the performance of AnyGPT on the Automatic Speech Recognition (ASR) task by calculating the Word Error Rate (WER) on the test-clean subsets of the LibriSpeech dataset~~\citep{panayotov2015librispeech}. 
We use Wav2vec 2.0 and Whisper Large V2 as baselines. Wav2vec 2.0 is pre-trained with 60,000 hours of speech and fine-tuned on LibriSpeech, while Whisper Large V2 is evaluated in a zero-shot setting but is trained with 680,000 hours of speech.
The results are shown in Table~\ref{tab:asr}.

\begin{table}[ht]
\centering\small
\begin{tabular}{lc}
\toprule
Method & WER $\downarrow$ \\
\midrule
Human-level~\citep{amodei2016deep} & 5.8 \\
Wav2vec 2.0~~\citep{Baevski2020wav2vec2A} & \textbf{{\color[HTML]{9B9B9B} 2.7}} \\
Whisper Large V2~~\citep{Radford2022RobustSR} & \textbf{2.7} \\
\midrule
AnyGPT & 8.5 \\
\bottomrule
\end{tabular}
\caption{Comparison results on ASR  task. We use  Word Error Rate (WER)  as the metric.}
\label{tab:asr}
\end{table}

\paragraph{TTS} We conduct a zero-shot Text-to-Speech~(TTS) evaluation on the VCTK dataset. 
The results are presented in Table~\ref{tab:tts}. 
We evaluate the TTS systems with speaker similarity and Word Error Rate (WER), where WER is focused on speech quality. More experimental details can be found in Appendix~\ref{sec:evaluation_appendix}.

\begin{table}[ht]
  \centering\small
  \begin{tabular}{lcc}
  \toprule
  Method & WER $\downarrow$ & SIM $\uparrow$ \\
  \midrule
  \textit{Ground Truth} & \textit{1.9} & \textit{0.93} \\
  VALL-E~~\citep{Wang2023NeuralCL} & 7.9 & 0.75 \\
  USLM~~\citep{zhang2023speechtokenizer} & \textbf{6.5} & \textbf{0.84} \\
  \midrule
  AnyGPT & 8.5 & 0.77 \\
  \bottomrule
  \end{tabular}
  \caption{Comparison results on  TTS task. }
  \label{tab:tts}
  \end{table}

\subsubsection{Music}

we evaluate AnyGPT's performance on the MusicCaps benchmark~~\citep{Agostinelli2023MusicLMGM} for both music understanding and generation tasks. We utilize the CLAP$_{score}$~~\citep{Wu2022LargeScaleCL, Huang2023MakeAnAudioTG} score as the objective metric, which measures the similarity between the generated music and a textual description. 
 
\begin{table}[h!]
\centering\small
\begin{tabular}{lc}
\toprule
Method                                           & CLAP$_{score}$  $\uparrow$\\
\midrule
\multicolumn{2}{l}{\textit{Music understanding}}        \\
\textless{}music, real caption\textgreater{}      & 0.16 \\
\textless{}music, generated caption\textgreater{} & 0.11 \\
\hline
\multicolumn{2}{l}{\textit{Music generation}}      \\
Riffusion~~\citep{Forsgren_Martiros_2022}                                        & 0.19 \\
Mousai~~\citep{Schneider2023MosaiTG}                                           & \textbf{0.23} \\
AnyGPT                                           & 0.14 \\
\bottomrule
\end{tabular}
\caption{Comparison results for music understanding and generation tasks. A metric scoring the alignment between music and textual captions is reported (CLAP$_{score}$). For music captioning, the CLAP$_{score}$ of  <music, real caption> pair and <music, generated caption> pair are computed for comparison.}
\label{tab:mu}
\end{table}
For the evaluation of music captioning, we found that existing objective metrics may be limited in expressing the performance in the music captioning task. The diversity and subjectivity of music lead to varying opinions from individuals. Only specific music genres and instruments possess distinctive characteristics that can be easily recognized. While recent studies~~\citep{Gardner2023LLarkAM} have explored this issue, it remains a challenging problem to address. 
To ensure an objective evaluation, we compute CLAP$_{score}$  of   <music, real caption> pairs and <music, generated caption> pairs for comparison.  These scores are averaged across the entire test set.

\subsection{Example Demonstrations }

After fine-tuning on the AnyInstruct-108k dataset, AnyGPT demonstrates the capability and potential in any-to-any multimodal dialogue.
We provide compelling conversation examples of AnyGPT in Appendix~\ref{sec:Examples-appendix}.  
These examples showcase AnyGPT is capable of comprehending and reasoning contents across various modalities in any combination.
Specifically, AnyGPT can comprehend instructions interwoven with multiple modalities, including text, voice, images, and music, and can adeptly select the appropriate multimodal combination for its reply.
The two-stage framework of semantic-acoustic hierarchical modeling empowers AnyGPT to generate voice responses that matches the timbre and emotion of a 3-second speech prompt.
For additional examples and to experience the speech and music content, we highly recommend visiting the \href{https://junzhan2000.github.io/AnyGPT.github.io/}{demo page}.


\section{Conclusion }

In this work, we introduced AnyGPT, an any-to-any multimodal language model that utilizes discrete representations for the unified processing of various modalities, including speech, text, images, and music. 
Discrete multimodal representations facilitate a seamless integration of new modalities—comparable to incorporating a foreign language—without necessitating alterations to the existing LLM architecture or training paradigms.
To equip the model to handle arbitrary combinations of multimodal inputs and outputs, we synthesize the first large-scale any-to-any multimodal instruction dataset, AnyInstruct-108k, consisting of multi-turn conversations that intricately interweave various modalities.
Experimental results indicate that AnyGPT achieves promising results in various cross-modal tasks and demonstrates that discrete representations can effectively and conveniently unify multiple modalities within a unified large language model. 

\section*{Limitations and Future Work}

\paragraph{Any-to-Any Multimodal LLM Benchmark}

The domain of any-to-any multimodal large language models (LLMs) is an emerging field of research. However, the lack of a dedicated benchmark to evaluate the models' capabilities across multiple dimensions, as well as to mitigate potential risks, presents a considerable challenge. 
Consequently, the development of a comprehensive benchmark is imperative.

\paragraph{Enhancing LLMs}
Although the multimodal LLMs with discrete representations can be trained stably, a higher loss is observed compared to unimodal training, preventing optimal performance in each modality.
Potential strategies to improve multimodal fusion could involve scaling LLMs and tokenizers or adopting a Mixture-Of-Experts (MOE) architecture to better manage diverse data and optimize performance.

\paragraph{Better Tokenizer}
In multimodal LLMs employing discrete representations, the tokenizer's quality sets a ceiling for the model's comprehension and generative potential. Enhancing the tokenizer can be approached from various angles, including the adoption of superior codebook training methods, the development of more cohesive multimodal representations, and the application of information disentanglement across various modalities.".

\paragraph{Longer Context}
Multimodal content, such as images and audio, often spans extensive sequences. AnyGPT, for instance, limits music modeling to 5 seconds, significantly restricting the practical usefulness of its audio output. Moreover, for any-to-any multimodal dialogue, an extended context allow for a higher number of conversational exchanges, thereby enriching the interaction's depth and complexity.

\newpage

\bibliography{colm2024_conference}
\bibliographystyle{colm2024_conference}
\appendix
\onecolumn


\section{pretraining}
\label{sec:pre-training}

\subsection{Data }

\begin{table*}[h!]
\centering\small
\renewcommand\arraystretch{1.25}
\resizebox{\textwidth}{!}{
\begin{tabular}{llcc}
\toprule
Modality                                                         & Dataset                  & Description                                                                                                                                                                & Sample Rate          \\
\midrule
\begin{tabular}[c]{@{}l@{}}Interleaved\\ Image-Text\end{tabular} & MMC4-core-ff             & \begin{tabular}[c]{@{}c@{}}101M image-interleaved documents collected from Common Crawl. \\ We use the mmc4-core split which is consist of 7.3M documents.\end{tabular} & 0.05                 \\
\midrule
\multirow{5}{*}{Image-Text}                                                       & Laion-2B                 & 2B image-text pairs from web.                                                                                                                                           & \multirow{5}{*}{0.3} \\
\cmidrule{2-3}
                                                                 & LAION-COCO               & 600M image-text pairs, where the caption is generated by BLIP.                                                                                                          &                      \\
                                                                 \cmidrule{2-3}
                                                                 & JourneyDB                & 4429K Midjourney images, with image caption.                                                                                                                            &                      \\
                                                                 \cmidrule{2-3}
                                                                 & LAION-Aesthetics         & Several collections of subsets from LAION 5B with high visual quality.                                                                                                  &                      \\
                                                                 \midrule
\multirow{5}{*}{Speech-Text}                                                     & Multilingual Librispeech & \begin{tabular}[c]{@{}c@{}}Processing audiobooks read from LibriVox, \\ we used a 44,000-hour subset of English.\end{tabular}                                           & 0.13                 \\
\cmidrule{2-4}
                                                                 & CommonVoice              & \begin{tabular}[c]{@{}c@{}}Microphone recordings from internet volunteers, \\ of which we used a 3000-hour subset of English.\end{tabular}                              & \multirow{3}{*}{0.27}                 \\
                                                                 \cmidrule{2-3}
                                                                 & GigaSpeech               & \begin{tabular}[c]{@{}c@{}}10,000 hours of English voice data sourced \\ from audiobooks, podcasts, and YouTube videos.\end{tabular}                                    &                      \\
                                                                 \midrule
\multirow{3.5}{*}{Music-Text}                                                             & Youtube-Music-1M         & 100M music-text pairs from Youtube.                                                                                                                                     & \multirow{3.5}{*}{0.25}                 \\
\cmidrule{2-3}
                                                                 & MusicGen-Synthesis       & \begin{tabular}[c]{@{}c@{}}20k music-text pairs extracted \\ from the AnyInstruct-108k dataset, synthesized by MusicGen.\end{tabular}                                    &               \\
                                                                 \bottomrule
\end{tabular}
}
\caption{Details of data used in pre-training stage.}
\label{tab:pdata}
\end{table*}

\label{sec:pre-training Data}

\subsection{pre-training}
\label{sec:Training Sample Construction}

We employ various templates to construct multimodal sentences, ensuring a diverse spectrum within our pre-training data. Each non-text modality content is identified by special tokens placed at both the beginning and end.
\begin{table*}[h]
\centering

\begin{tabular}{lcc}
\toprule
                                                                          & Pre-training Stage & Fine-tuning Stage        \\
                                                                          \midrule
\begin{tabular}[c]{@{}l@{}}Gradient clipping\\ (Global-norm)\end{tabular} & 1.0 &1.0                      \\
Batch size                                                                & 480               &     64                     \\
Max   length                                                              & 4500              &      4500                    \\

Training steps                                                            & 81000             & 5000 \\
Learning rate scheduler                                                   & cosine & cosine                \\
Peak learning rate                                                        & 6e-5              & 2e-5 \\
Warmup ratio                                                              & 0.03 & 0.03                   \\
Optimizer                                                                 & Adam & Adam     \\
GPU                                                                       &  A100 & A100     \\
\bottomrule
\end{tabular}
\caption{Training hyperparameters used in experiments.}
\label{tab:HYPE}
\end{table*}
Typically, the paired data comprises a non-text modality (X) - such as images, speech, or music - and its corresponding text, which could be a caption or transcription. We prompt OpenAI GPT-4 to generate hundreds of bidirectional instructions, specifically X-to-text or text-to-X such as "Please generate an image based on the provided text."
Given a token sequence (S) and related text (T), we randomly pick a generation direction alongside an instruction (I) from our pre-established pool, forming a triplet (I, S, T). This triplet is then incorporated into a sequence using the template \\
 \textbf{[Human]:} \{\textit{I}\}.\{\textit{S}\}\textbf{<eoh>.} 
\textbf{[AnyGPT]:} \{\textit{T}\}\textbf{<eos>.} or its variant 
\\\textbf{[Human]:} \{\textit{I}\}.\ This is input:\{\textit{T}\}\textbf{<eoh>.} 
\textbf{[AnyGPT]:} \{\textit{S}\}\textbf{<eos>.}, depending on the generation direction.

For interleaved multimodal data, like a web document with interspersed images and text, we directly replace non-text content with the corresponding tokens sequence as they naturally form sentences.

As most of the image and music data are sourced from the web, there is a certain level of noise that can affect the quality of multimodal generation. Consequently, after the initial pre-training, we selectively utilized high-quality datasets—JourneyDB and LAION-Aesthetics for text-to-image generation, and LAION-COCO for image captioning. For music data, we incorporated the AnyInstruct-108k dataset. The remaining data were kept unchanged, and we continued to pre-train the model for an additional 4000 steps.

We report the detailed training hyperparameters of AnyGPT in Tab~\ref{tab:HYPE}.

\section{Instruction Tuning}
\label{sec:Instruction Tuning-appendix}

\begin{figure*}[h] 
  \centering
  \includegraphics[width=0.85\textwidth]{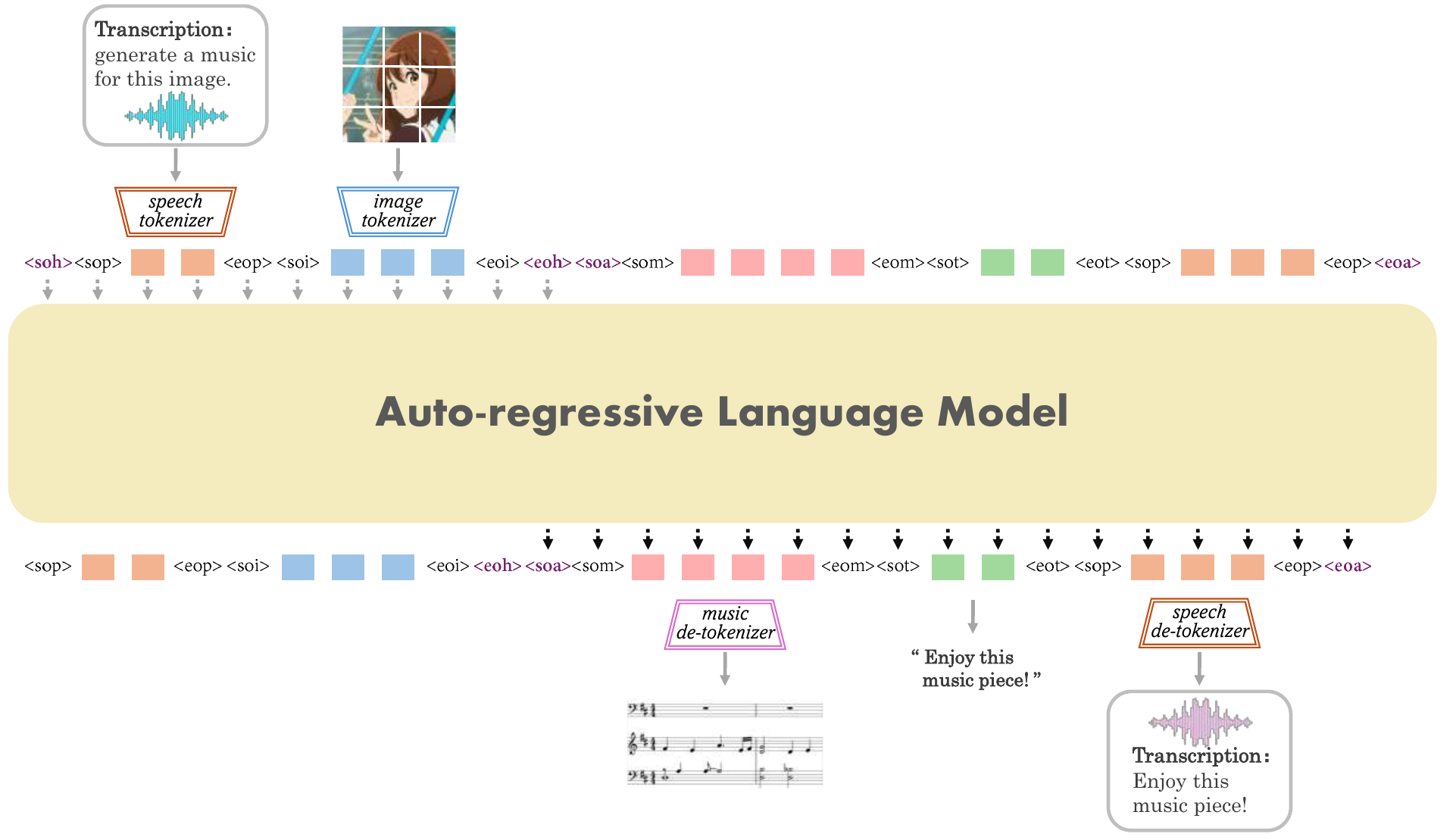}
  \caption{An example of an AnyGPT multimodal dialogue: the input is an image and a voice command to generate music. The output is music that meets the requirements, along with corresponding text and voice responses. All data are processed into discrete tokens and are autoregressively processed by the LLM.}
  \label{fig:model2}
  \end{figure*}

\section{Evaluation}
\label{sec:evaluation_appendix}

\begin{table}[h!]
\centering
\begin{tabular}{l|cccc}
\toprule
Generated Modality & Text                 & Image    & Speech   & Music    \\
\midrule
Decoding Strategy  & Beam Search & Sampling & Sampling & Sampling \\
Beam size          & 5                    & -        & -        & -        \\
Top-P              & -                    & 0.7      & 0.7      & 1.0      \\
Repetition Penalty & 1.0                  & 1.0      & 1.0      & 1.15    \\
\bottomrule
\end{tabular}
\caption{Details of generation decoding strategies used in evaluation.}
\label{tab:gene}
\end{table} 
We conduct a zero-shot Text-to-Speech~(TTS) evaluation on the VCTK dataset. There is no overlap in speakers between our training data and the VCTK dataset. We randomly select a  3-second clip from each speaker as the vocal prompt along with a separate text as input.

The results can be found in Table~\ref{tab:tts}. We evaluate the TTS systems with speaker similarity and WER. To evaluate the speaker similarity between the generated speech and the prompt speech, we employ WavLM-TDNN\footnote{\url{https://github.com/yangdongchao/UniAudio/blob/main/UniAudio/tools/evaluation/compute_similarity_vc.py}}. It can generate speaker embeddings for both the generated speech and the prompt speech, then compute the cosine similarity between these embeddings. WER is calculated using the Whisper medium model to transcribe the generated speech, with lower WER indicating higher quality of the synthesized speech.

We compare our model with  VALL-E and USLM, both of which employ two autoregressive models for speech modeling. They utilize Encodec and SpeechTokenizer, respectively, as speech tokenizers.

\vspace{-6mm}
\section{Prompts for Constructing Multimodal Interleaved Instruction Data}
\label{sec:anyinstrurct-prompts-appendix}
\vspace{-4mm}
In the first stage of our pipeline to construct multimodal interleaved instruction data (Sec.~\ref{sec: mm-interleaved-data}) with GPT4. 
To facilitate reproducibility, we detail our prompts to the language model for brainstorming a topic pool (Fig.~\ref{fig:prompts4topic}), constructing chatting scenarios (Fig.~\ref{fig:prompts4scenario}), and detailing the chat contents (Fig.~\ref{fig:prompts4chat}), with multimodal content written as their text descriptions.

\begin{center}
\vspace{-2mm}
\noindent\begin{minipage}{\textwidth}
\begin{tcolorbox}[colback=gray!10,colframe=black,width=\textwidth,arc=2mm,boxrule=0.5pt]
\textbf{Prompt:} Please list me 50 **non-academic** conversation topics about \texttt{\{metatopic\}} between an ordinary person and a helpful chatbot. Each topic should be made up of 1-10 words and the conversation contain understanding and generation of images or music.\\
~\\
\textbf{GPT4:} \texttt{\{50 sampled topics\}}\\
~\\
\textbf{Prompt:} continue\\
~\\
\textbf{GPT4:} \texttt{\{50 more sampled topics\}}\\
~\\
$\cdots\cdots$
\end{tcolorbox}
\vspace{-3mm}
\captionof{figure}{Prompts for brainstorming chat topics. We prepare 100 \texttt{metatopic} and repeat the conversation to brainstorm topics for 4 rounds for each of the metatopic. This gives 200 topics per metatopic and a total of 20,000 topics in our final topic pool.}\label{fig:prompts4topic}
\end{minipage}
\noindent\begin{minipage}{\textwidth}
\vspace{5mm}
\begin{tcolorbox}[colback=gray!10,colframe=black,width=\textwidth,arc=2mm,boxrule=0.5pt]
\textbf{Prompt:}\\
You are a creative assistant. I am now asking you to help me brainstorm some chatting scenarios where the user asks the agent for help. \\
Note that the scenarios should be between ordinary people and a helpful chatbot, and it should not be an academic discussion!\\
During the conversation, the speakers can use images or music to help convey information (but do not use video!). And the user should ask questions about it if he/she provides an image or a piece of music.\\
Note that the image should not be a chart.\\
Note that the images and music should not be the famous masterpieces that may arise copyright issues.\\
~\\
Here are some of my ideas, and please show me more in the same format as mine.\\
~\\
\texttt{\{demonstrations\}}\\
~\\
Here's the topics for you to try: \texttt{\{topics\}}\\
~\\
Now it's your turn. In these scenarios, \texttt{\{requirements\}}.\\
~\\
\textbf{GPT4:}\\
\texttt{\{synthetic scenarios of the provided topics, following requirements\}}
\end{tcolorbox}
\vspace{-2.5mm}
\captionof{figure}{Prompts for constructing chat scenarios. In each API call, we sample 5 different \texttt{\{demonstrations\}}, with each containing a topic and detailed description of the scenarios. And we sample 10 different \texttt{\{topics\}} for GPT4 to synthesize scenarios. To ensure the diversity of user and chatbot actions, we explicitly sample \texttt{\{requirements\}} from ``the user provide images'', ``the user share music'', ``the user asks for music'', and ``the user asks for images''. We up weight ``the user share music'' as we observe that the model tends to omit this requirement.}\label{fig:prompts4scenario}
\end{minipage}

\noindent\begin{minipage}{\textwidth}
\begin{tcolorbox}[colback=gray!10,colframe=black,width=\textwidth,arc=2mm,boxrule=0.5pt]
\textbf{Prompt:}\\
You are helping me to write conversations about a user talking to a chatbot named AnyGPT. \\
~\\
In the conversations, both the user can provide images or music to help express her/his needs and ideas. And the chatbot AnyGPT can also respond to the user with images or music in its utterances.\\
~\\
The images and music in the chat are in the form of image descriptions and music descriptions like \underline{[image: description]} and \underline{[music: description]}, respectively. The user should provide images and music in this format and the chatbot will respond to the user like this as well.\\
~\\
Note that at most one music appears in one conversation and the description of music should be straightforward, focusing on genres and instruments, and never mention a known music directly. \\
~\\
Before each conversation, I will first show you a scenario and you can start writing about the chat.\\ 
~\\
Here is an example:\\
---\\
\texttt{\{demonstrations\}}\\
---\\
~\\
Now it's your turn for the next conversation. You only need to answer following the format in which the user and AnyGPT take turns. \\
The conversation should be consistent with the introduction to the scenario.\\
Remember that the utterances should be concise, try to use 5-15 words per utterance. \\
Note that: the user utterance should always be a question or instruction.\\
In some turns, the user provides an image or a piece of music and asks a question or makes an instruction to AnyGPT relating to the provided image or music.\\
In other turns, the user requests AnyGPT to generate the required images or music. \\
Note that: the description of music should focus on genres, style, and instruments. And make the description of images and music within [image: ] or [music: ] more detailed.\\
Note that: never directly include a famous person's name in the image descriptions or mention a piece of known music in the music description.\\
Tips: when the user asks to convert between images and music, AnyGPT should first utter his understanding of the input image or music before generating the requested result.\\
Keep the dialog in 2 or 3 rounds.\\
Each dialog should include one music and at most 2 images.\\
~\\
---\\
In this conversation, \texttt{\{new\_scenario\_description\}}\\
~\\
\textbf{GPT4:}\\
\texttt{\{A synthetic chat according to the scenario description.\}}
\end{tcolorbox}
\captionof{figure}{Prompts for writing chat content. For each API call, we sample 3 demonstrations. Each demonstration contains a scenario description, as the \texttt{\{new\_scenario\_description\}}, and the corresponding chat. }\label{fig:prompts4chat}
\end{minipage}
\end{center}

\newpage{}

\section{Examples Demonstration}
\label{sec:Examples-appendix}
\begin{figure*}[h] 
\centering
\includegraphics[width=1.0\textwidth]{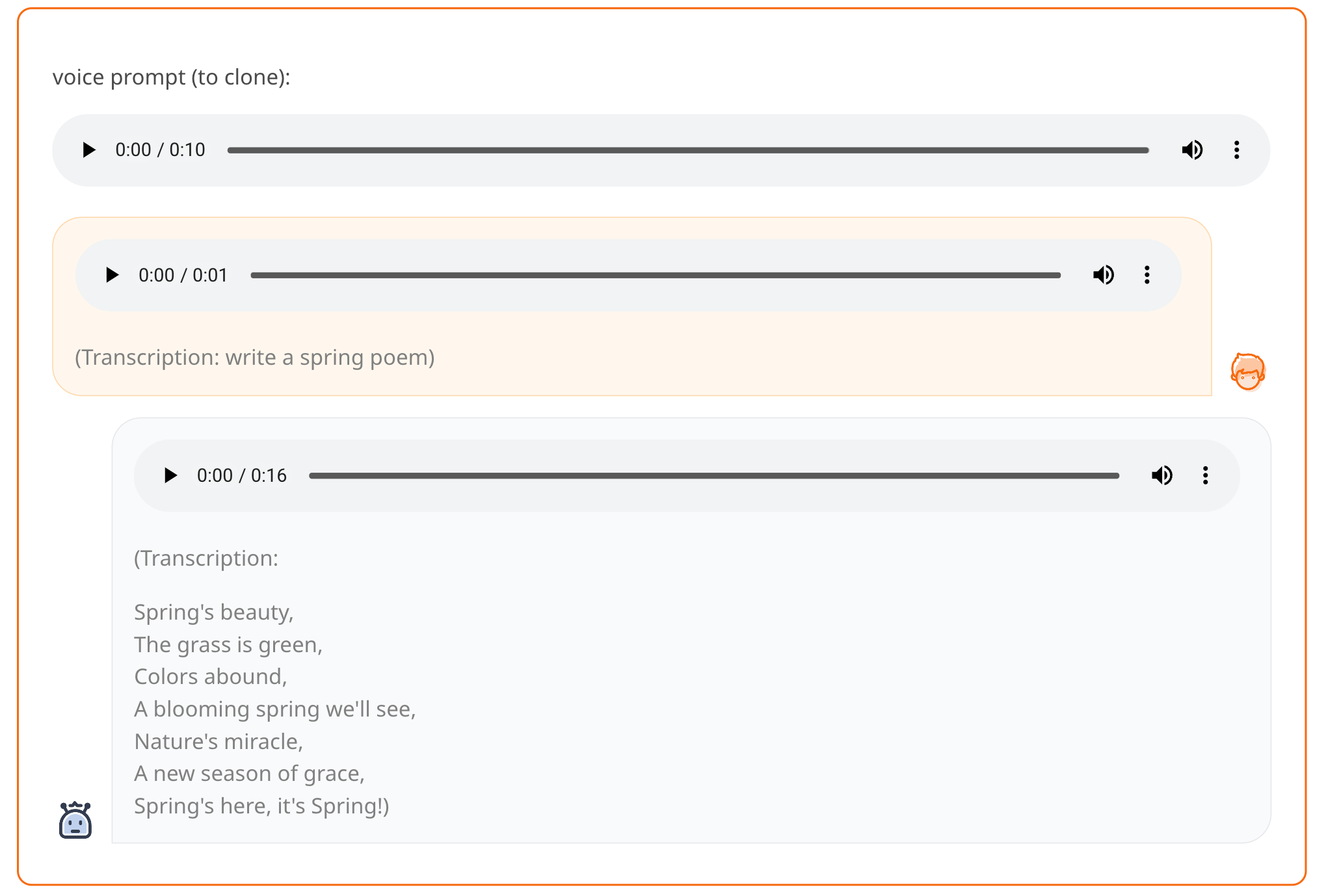}
\caption{Speech conversations (Voice Clone)}
\label{fig:example1}
\end{figure*}

\begin{figure*}[h] 
\centering
\includegraphics[width=1.0\textwidth]{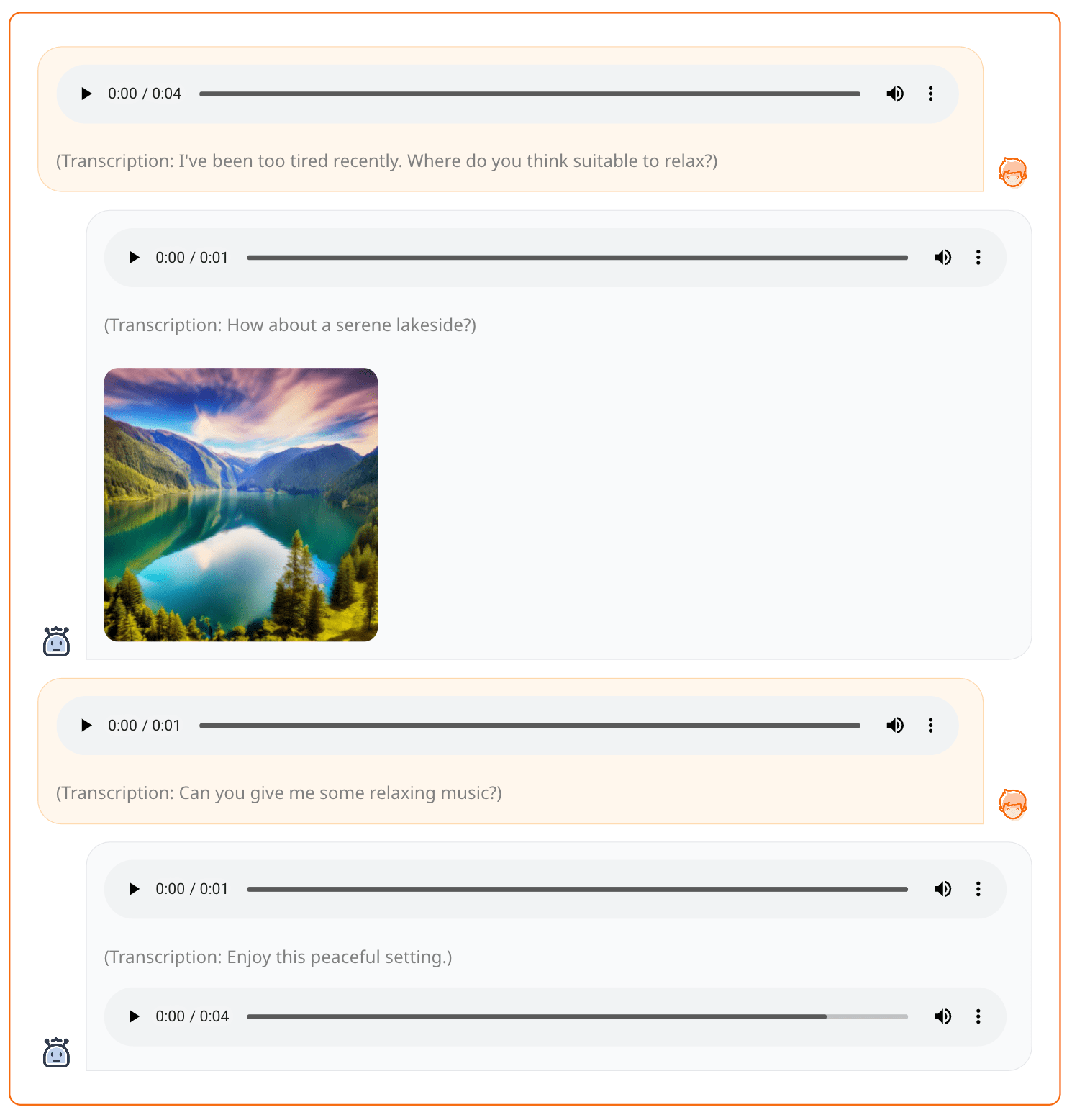}
\caption{Speech Instruction + Image → Text + Music + Speech Response}
\label{fig:example2}
\end{figure*}

\begin{figure*}[h] 
\centering
\includegraphics[width=1.0\textwidth]{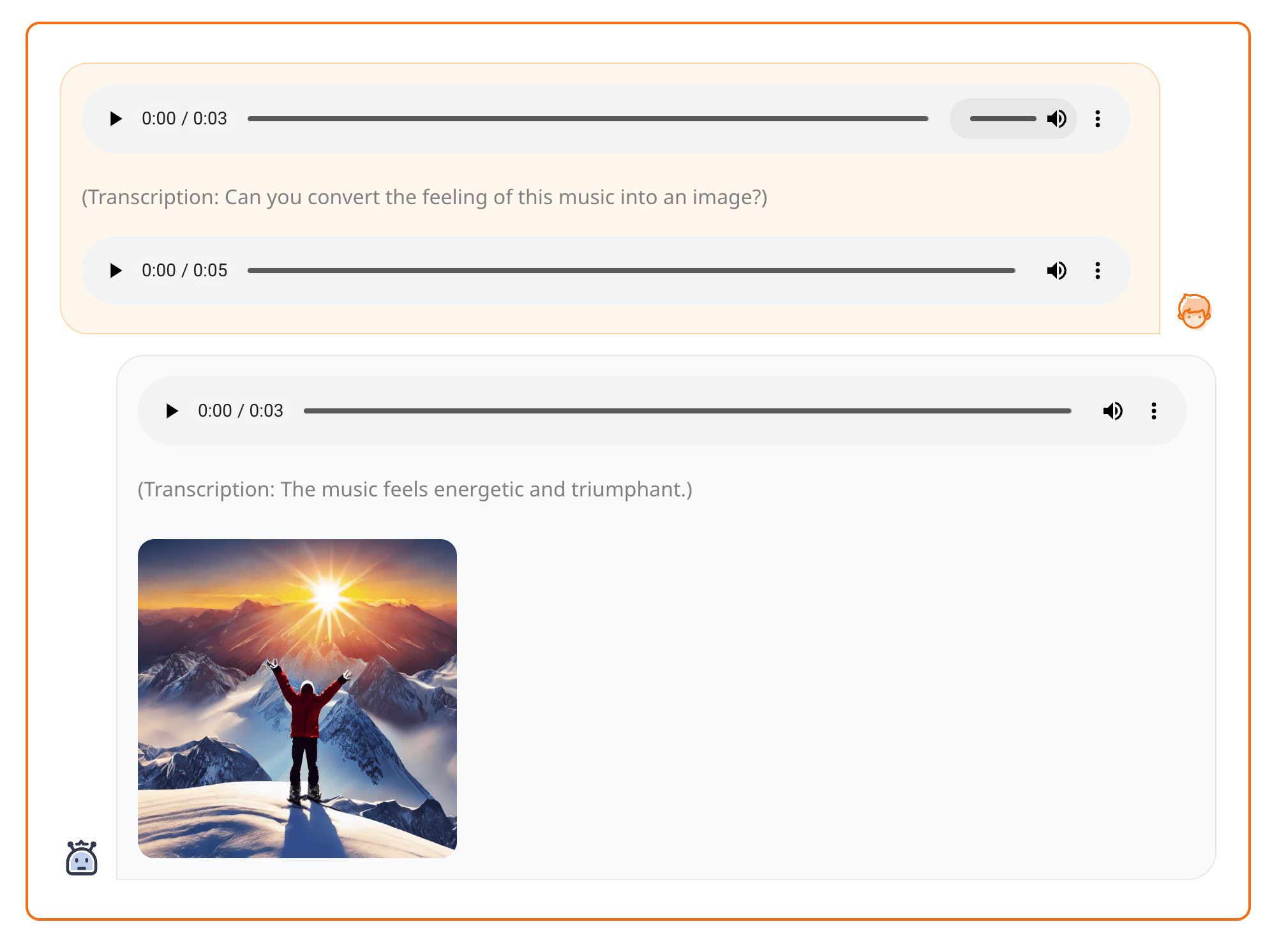}
\caption{Speech Instruction + Music → text + Music + Speech Response}
\label{fig:example3}
\end{figure*}

\begin{figure*}[h] 
\centering
\includegraphics[width=1.0\textwidth]{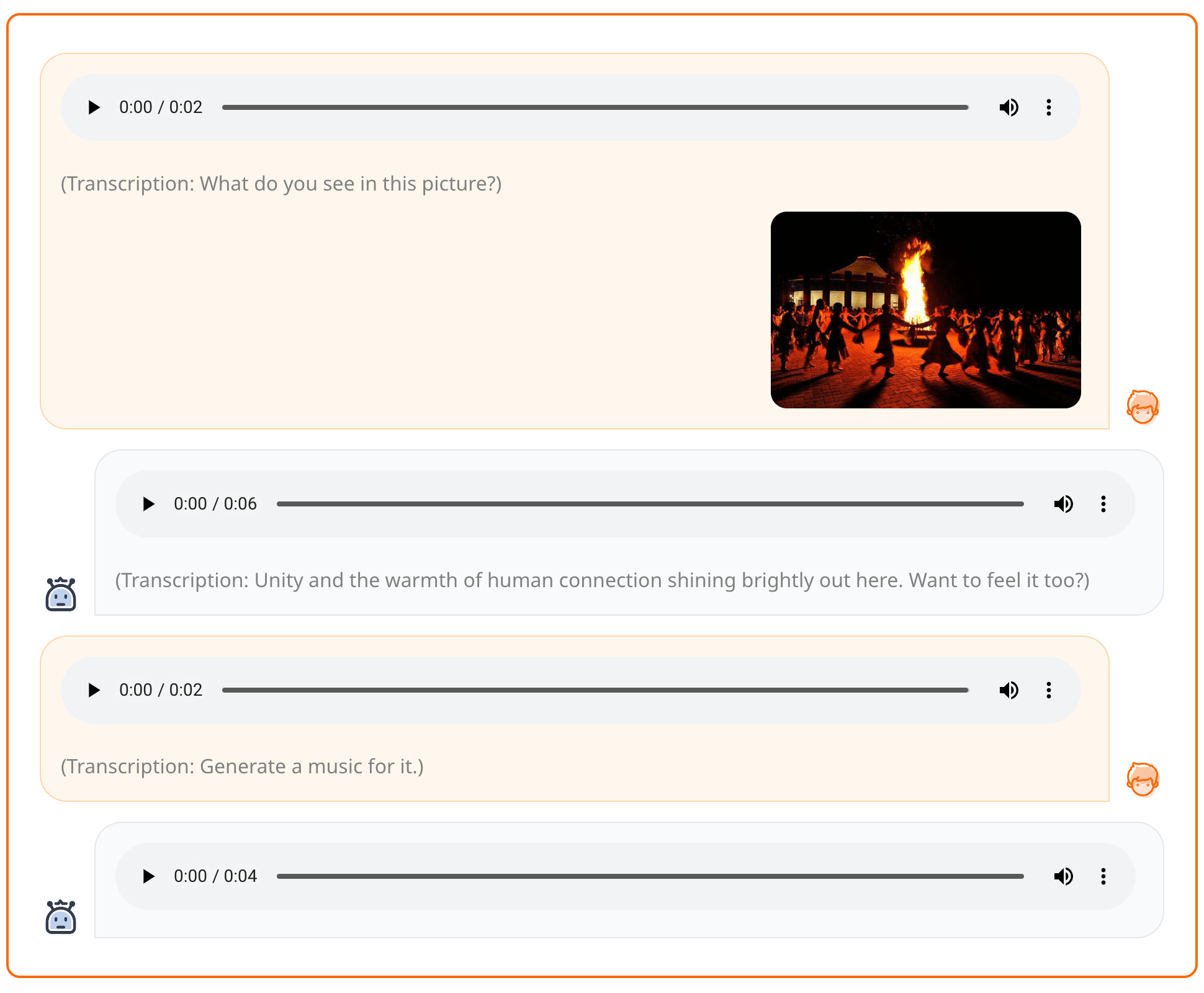}
\caption{Speech Instruction + Image → text + Music + Speech Response}
\label{fig:example4}
\end{figure*}

\begin{figure*}[h] 
\centering
\includegraphics[width=1.0\textwidth]{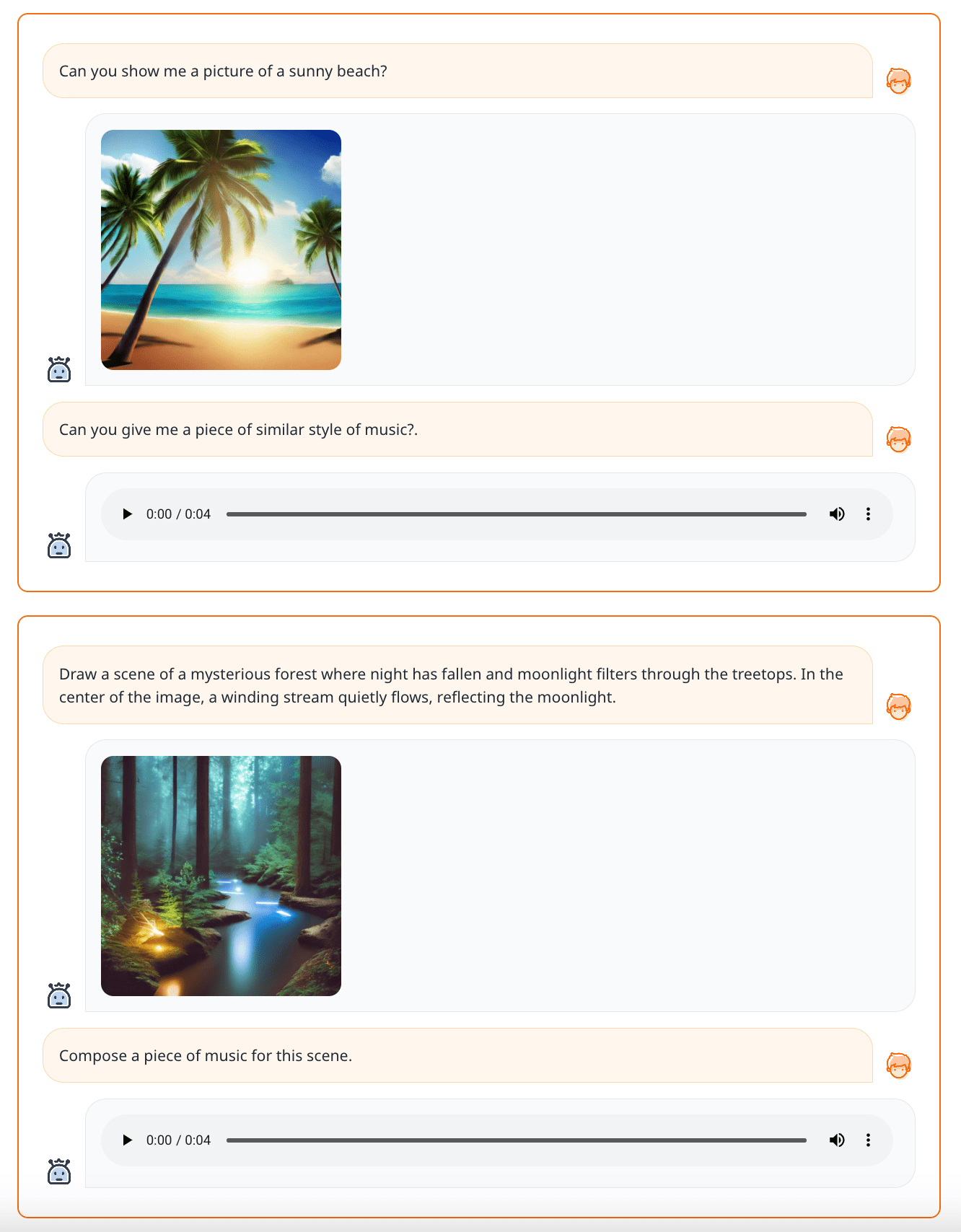}
\caption{Text → Image + Music}
\label{fig:example5}
\end{figure*}

\begin{figure*}[h] 
\centering
\includegraphics[width=1.0\textwidth]{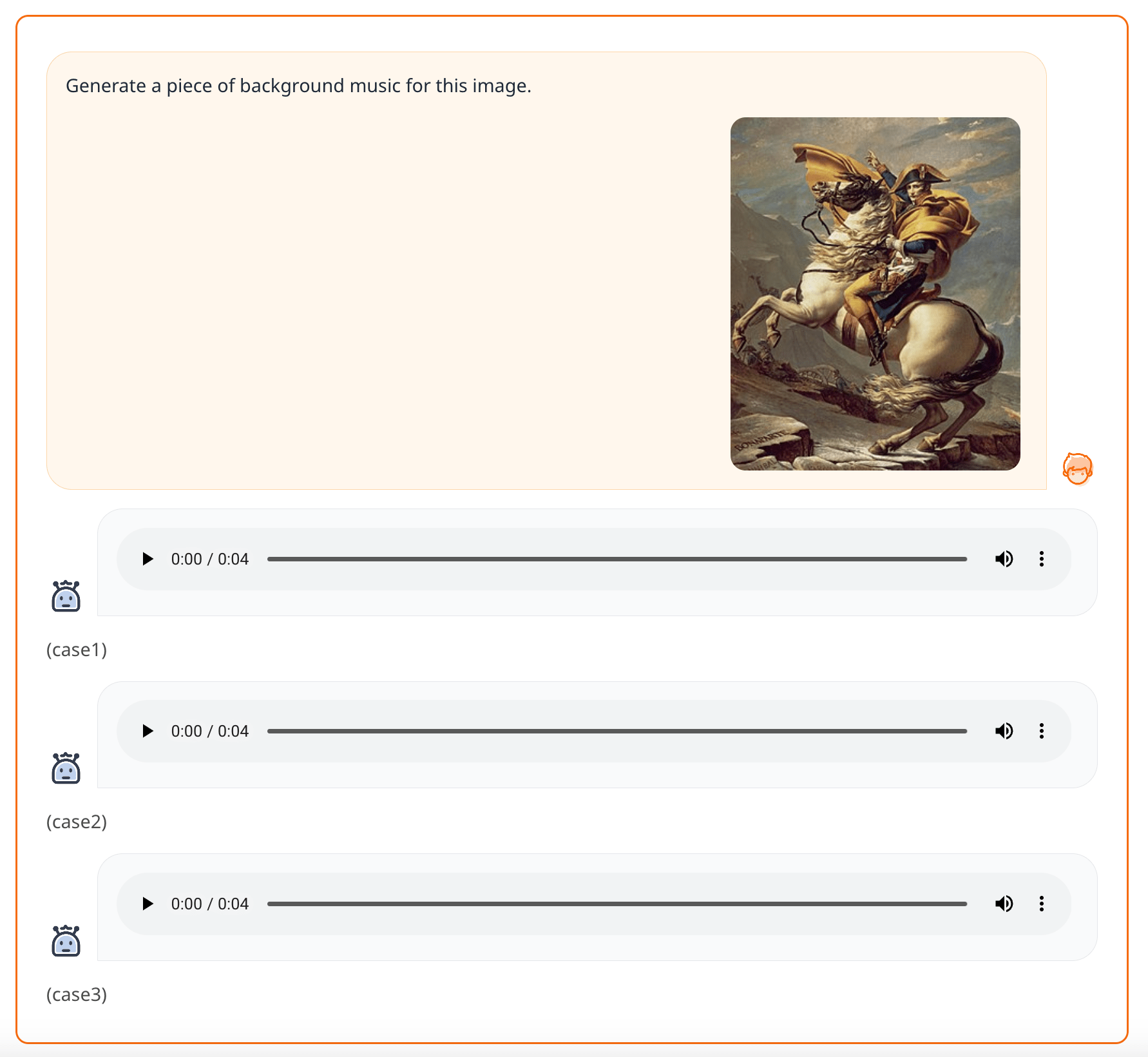}
\caption{Text + Image → Music}
\label{fig:example6}
\end{figure*}

\begin{figure*}[h] 
\centering
\includegraphics[width=1.0\textwidth]{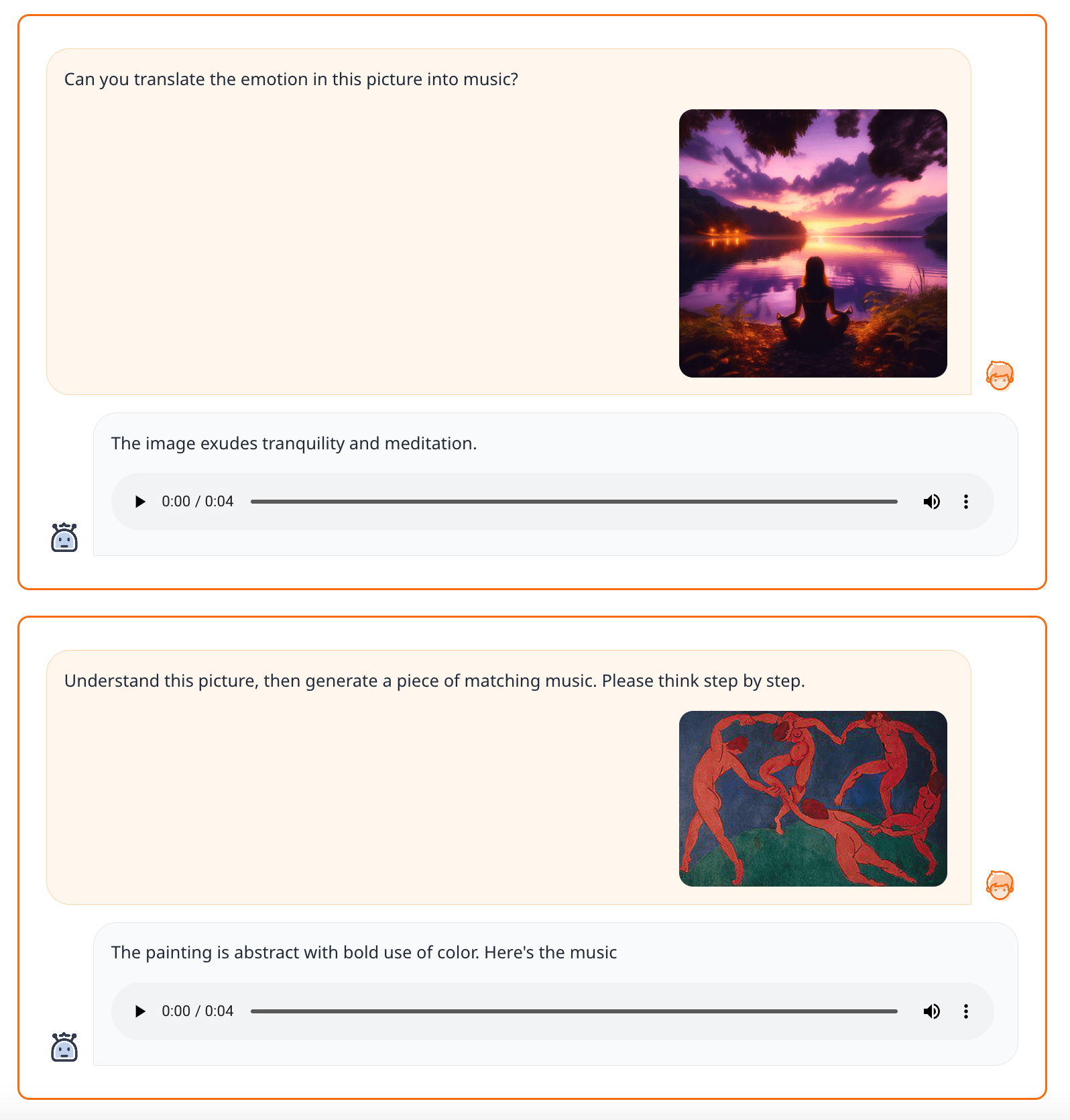}
\caption{Text + Image → Text + Music}
\label{fig:example7}
\end{figure*}

\begin{figure*}[h] 
\centering
\includegraphics[width=1.0\textwidth]{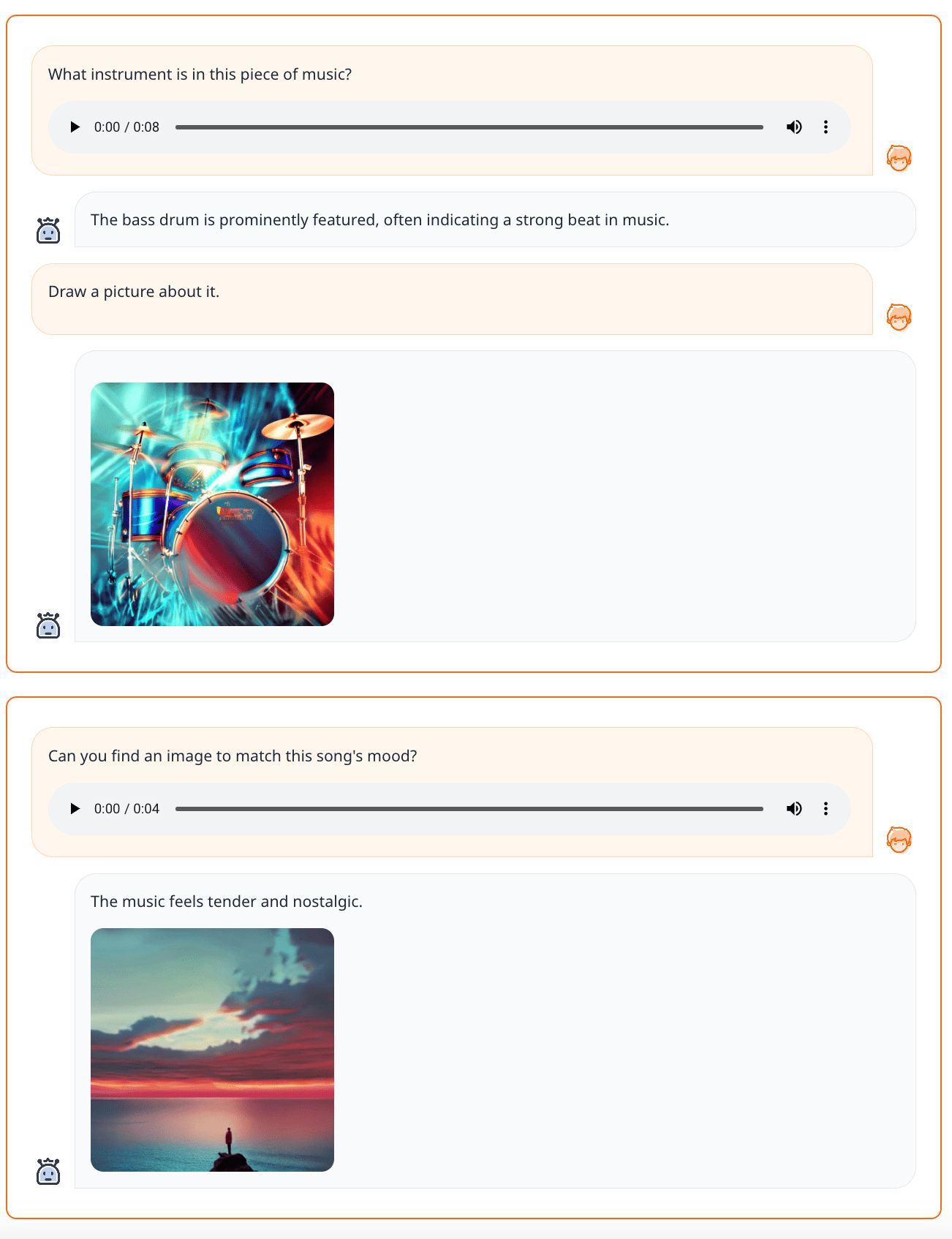}
\caption{Text + Music → Text + Image}
\label{fig:example8}
\end{figure*}

\begin{figure*}[h] 
\centering
\includegraphics[width=1.0\textwidth]{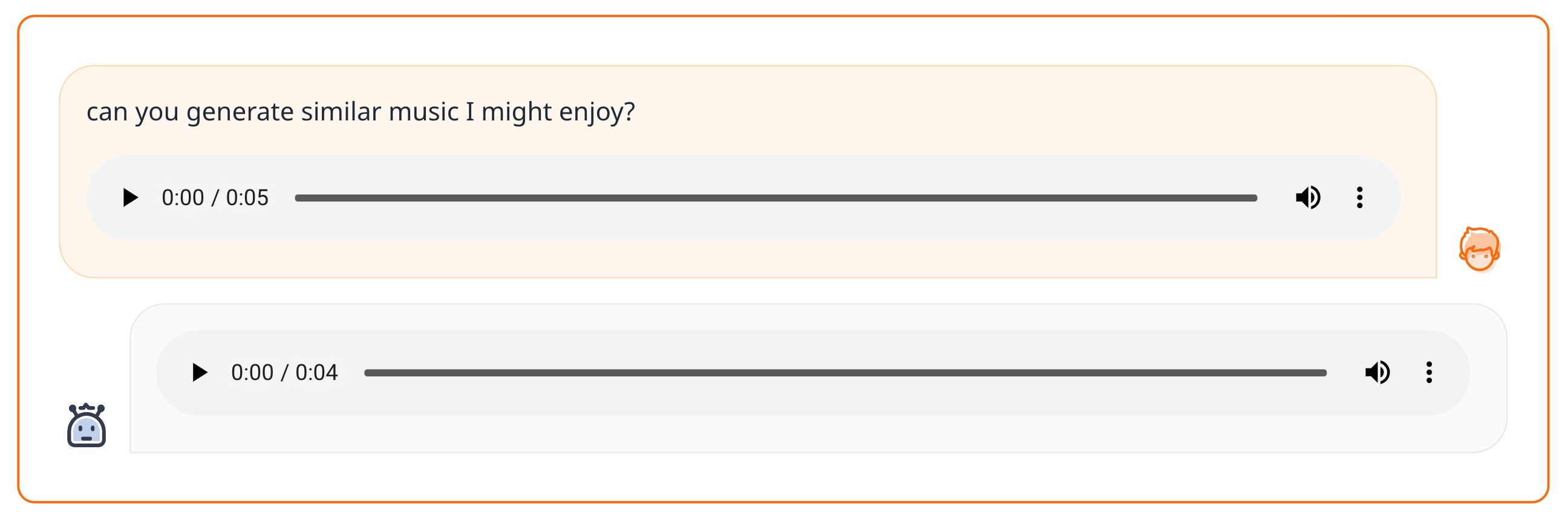}
\caption{Text + Music → Muisc}
\label{fig:example9}
\end{figure*}

\end{document}